\documentclass[letterpaper]{article} 
\usepackage[submission]{aaai23}  
\usepackage{times}  
\usepackage{helvet}  
\usepackage{courier}  
\usepackage[hyphens]{url}  
\usepackage{graphicx} 
\usepackage{natbib}  
\usepackage{caption} 
\usepackage{algorithm}
\usepackage{algorithmic}

\usepackage{newfloat}
\usepackage{listings}
\DeclareCaptionStyle{ruled}{labelfont=normalfont,labelsep=colon,strut=off} 
\floatstyle{ruled}
\newfloat{listing}{tb}{lst}{}
\floatname{listing}{Listing}
\title{Planning with Complex Data Types in PDDL}
\author {
    Mojtaba Elahi, 
    Jussi Rintanen 
}
\affiliations {
    Aalto University\\
    mojtaba.elahi@aalto.fi, jussi.rintanen@aalto.fi
}

\usepackage{amsmath}
\usepackage{booktabs}
\usepackage{mathabx}
\usepackage{scalerel}
\usepackage{multirow}

\newtheorem{example}{Example}

\newcommand{\hs}{\hspace{-0.1em}}
\newcommand{\eff}{e\hs\hs{f}\hs\hs\hs{f}}
\newcommand{\args}{ar\hs{g}\hs{s}}
\newcommand{\pre}{pr\hs{e}}

\newcommand{\tpl}[1]{\langle {#1} \rangle}
\newcommand{\tuple}[1]{\left\langle {#1} \right\rangle}
\newcommand{\Tuple}[1]{%
  \vcenter{%
    \hbox{%
      $\displaystyle\stretchleftright
        {\bigg\langle} {\substack{#1}} {\bigg\rangle}$%
    }%
  }%
}

\makeatletter
\newcommand{\subalign}[1]{%
  \vcenter{%
    \Let@ \restore@math@cr \default@tag
    \baselineskip\fontdimen10 \scriptfont\tw@
    \advance\baselineskip\fontdimen12 \scriptfont\tw@
    \lineskip\thr@@\fontdimen8 \scriptfont\thr@@
    \lineskiplimit\lineskip
    \ialign{\hfil$\m@th\scriptstyle##$&$\m@th\scriptstyle{}##$\hfil\crcr
      #1\crcr
    }%
  }%
}
\makeatother

\newcommand{\mySubstack}[1]{%
  \let\scriptstyle\textstyle
  \begin{subarray}{l}
    \\[1pt]
    #1
    \\[1pt]
  \end{subarray}
}

\newcommand{\fix}[2]{\text{fix}\left(\mySubstack{#1,\\[1pt] #2 \\}\right)}
\newcommand{\dom}[1]{\text{dom}({#1})}
\newcommand{\Dom}[1]{\text{dom}\left({#1}\right)}

\newcommand{\semantics}[2]{\left\ldbrack {#1}\right\rdbrack^{#2}}
\newcommand{\rsemantics}[2]{{}^r\ldbrack #1 \rdbrack^{#2}}

\begin{document}

\maketitle

\begin{abstract}
Practically all of the planning research is limited to states 
represented in terms of Boolean and numeric state variables. Many
practical problems, for example, planning inside complex software
systems, require far more complex data types, and even real-world 
planning in many cases requires concepts such as sets of objects, which
are not convenient to express in modeling languages with scalar types only.

In this work, we investigate a modeling language for complex software
systems, which supports complex data types such as sets, arrays,
records, and unions. We give a reduction of a broad range of complex
data types and their operations to Boolean logic, and then map this
representation further to PDDL to be used with domain-independent PDDL
planners. We evaluate the practicality of this approach, and provide
solutions to some of the issues that arise in the PDDL translation.
\end{abstract}

\section{Introduction}

PDDL is the leading specification language used by the AI planning community
in expressing (classical) planning problems and solving them by
domain-independent planners.
Most of the planners for the classical planning problem (with a unique initial
state and actions that are deterministic) implement Boolean state variables only.
Some additionally support numerical state variables, integers or reals.
While this is theoretically sufficient to express a broad class of
complex planning problems, the practical limits of PDDL are sometimes
encountered. This is particularly evident when PDDL specification are not
written by hand, and need to be generated by programs written in conventional
programming languages. There are numerous examples of this both in standard
PDDL benchmark problem sets as well as in the planning literature.
PDDL is in these cases used an intermediate language to interface with
domain-dependent planners.

The idea of more powerful specification language, allowing the easier
expression of complex planning problems, and to reduce the need for
ad hoc problem generators, is not new.
A prominent example is {\em Functional STRIPS} \cite{Geffner00}.
If PDDL's predicates $P(t_1,\ldots,t_n)$ are viewed as {\em arrays} as in
programming languages, PDDL limits to arrays indexed by object names
or by actions' parameter variables only, and only Boolean values as elements.
Functional STRIPS goes further, allowing more complex array indexing,
including with object-valued expressions and nesting of array indexing.
Specifications in this kind of language can often outperform PDDL in elegance and succinctness.

The goal of this work is, similarly to Functional STRIPS, to increase
the expressivity of specification languages. We believe that
a planning specification language should support a broad collection of
different data types, including scalars like Booleans, numeric types, and enumerated types,
as well as compound types such as records, unions, arrays, sets and lists.
This has been motivated by our work on using existing planning technology
in the creation of intelligent software systems, which handle complex
structured and relational data, as found in almost any software application.

In this work, we will first present an expressive language that supports several
complex data types, and then we will give reductions of that language to
an intermediate Boolean representation.
This representation can be used as a basis of implementations in different types
of intermediate and lower level modeling languages.

As an obvious language to reduce the Booleanized representation to is PDDL \cite{PDDL98},
due to the existence of dozens of implementations of PDDL in scalable, robust planners.
The front-end of our planning system uses the extended specification language that supports complex
datatypes. The expressions in this language, and the actions based on them, are
reduced to a purely Boolean representation. The final step is the generation of PDDL, so that
existing planners can be used for finding plans.

Intuitively it is clear that reducing complex data types to PDDL is possible,
but it is not clear how practical those reductions are, and how efficiently
existing planners for PDDL can solve the resulting PDDL specifications.
Many existing planners support a large fragment of PDDL, but may translate some parts
of PDDL representations to specific normal forms, and these normal form translations
may increase the size of the representations. For instance, planner back-ends may require
actions' preconditions to be conjunctions of (positive) literals.
For this reason, to support many state-of-the-art planners better, we develop techniques to further
process the PDDL to forms that are better digestible by existing planners.

In the experimental part of the work, we try out the resulting planner front-end
with different state-of-the-art PDDL planners as back-ends.
We demonstrate their ability to solve complex problems that would be tedious
and unintuitive to express in PDDL.

The experiments help identify bottle-necks in existing planners, which could
aid in developing existing planning technology to better handle more complex
problem specifications.

\section{Target Planning Model}
We define a planning problem by a quadruple of $P = \tuple{\mathcal{V}, s_0, g, \mathcal{A}}$, where $\mathcal{V}$ is the set of state variables, $s_0$ defines the initial state by specifying the values of all state variables in $\mathcal{V}$, $g$ is a Boolean formula describing the set of goal states (i.e., a state $s$ is a goal state iff $g$ holds in $s$), and $\mathcal{A}$ denotes the set of actions. An action $a \in \mathcal{A}$ is defined by a triple of $a = \tuple{\args_a, \pre_a, \eff_a}$. $\args_a$ specifies the parameters of the action. A parameter is a variable defined in the scope of the action. $\pre_a$ is the precondition of the action, which is a Boolean formula specifying the applicability of the action in a state. $\eff_a$ is a set of conditional assignments defining how a state will be modified after applying the action. A conditional assignment is a pair $\tuple{cond, r := e}$, where $cond$ is a Boolean formula defining the condition and $r := e$ is the assignment that assign the value of $e$ to the reference variable $r$ after executing action $a$ if the condition $cond$ holds.

Theoretically, variables of $\mathcal{V}$ could be of any type, but in this work we limit to Boolean state variables, which is the most commonly used type in planners that support PDDL.

\subsection{Extended Language}
The goal of our extended language is to support more complex data types and standard operations on them. The data types we are covering are described in Table \ref{tab:datatypes}.

\begin{table}
\centering
\begin{tabular}{ll}
    \toprule
    Notation                         & Data type \\
    \midrule
    $bool$                           & Boolean \\
    $n..m$                           & bounded integer \\
    $\{c_1, \ldots, c_n\}$           & enumerated type with items \\
                                     & $c_1, \ldots, c_n$ \\

    $\{t\}$                          & set of elements of type $t$ \\
    $t_v\{t_k\}$                     & associative array with index \\
                                     & type of $t_k$ and value type of $t_v$ \\
    $\tuple{t_1, \ldots, t_n}$       & $n$-tuple with elements of types \\
                                     & $t_1, \ldots, t_n$ \\
    $\{f_1: t_1, \ldots, f_n: t_n\}$ & record with $n$ fields \\
    $[u_1: t_1, \ldots, u_n: t_n]$   & union with $n$ components \\
    \bottomrule
\end{tabular}
\caption{Data types}
\label{tab:datatypes}
\end{table}

Using standard operations over the variables of these data types makes it possible to compactly formulate arbitrary nested expressions to define complex actions and goal condition. The operations we support in the extended language are listed in Table \ref{tab:operations}.

\begin{table}
\centering
\begin{tabular}{ll}
    \toprule
    Data type      & Operations \\
    \midrule
    Boolean        & $\neg, \wedge, \vee$ \\
    bounded integer & $+, -, \times, \div, <, \leq, =, \neq, \geq, >$ \\
    enum           & $=$ \\
    set            & $\in, \subseteq, \cup, \cap, \setminus$ \\
    array          & index access \\
    tuple          & element access \\
    record         & field access \\
    union          & component access \\
    \bottomrule
\end{tabular}
\caption{Supported operations for each data type}
\label{tab:operations}
\end{table}

The intermediate representations of values of all supported data types are
listed in Table \ref{tab:representations} (the $\text{dom}(t)$ function used in this table is defined in Table \ref{tab:helper_functions}).
Expressions that represent values of these data types have the same structure, except
that what is the atomic part of the representation will be replaced by general propositional
formula.
For example, the Booleanized representation of numeric expressions of type $n..m$
is as $m-n+1$ element vectors of propositional formulas.

\begin{table}
\centering
\begin{tabular}{ll}
\toprule
Data type               & Representation \\
\midrule
bool                    & Boolean variable \\
$n..m$                  & vector of $m-n+1$ Boolean variables \\
$\{c_1,\ldots,c_n\}$     & vector of $n$ Boolean variables \\
$\tuple{t_1,\ldots,t_n}$ & $n$-tuple of representations of $t_1,\ldots,t_n$ \\
$\{t\}$                 & vector of $\text{dom}(t)$ Boolean variables\\
$t_v\{t_k\}$            & vector of $\text{dom}(t_k)$ representations of $t_v$ \\
\end{tabular}
\caption{Representation of values}\label{tab:representations}
\end{table}

\section {Reduction to Boolean Representation}

Although versions of PDDL support numeric and object fluents \cite{kovacs11} (which are similar to integer and enumerated data types), the lack of widespread support of these features by planners made us choose the version of PDDL as our target language that only supports Boolean state variables.
Consequently, having only Boolean state variables means we need to represent state variables of different data types with Boolean variables and translate all of their expressions into Boolean formulas.

\subsection{Scalar Data Types}

Representation of Boolean state variables as individual Boolean variables is trivial. Other variables need to be represented by collections of Booleans.
Here we only consider a {\em unary} representation for bounded integers and Enums. In other words, we represent a bounded integer $n..m$ as a vector of $m-n+1$ Boolean variables; also, an Enum type variable with $n$ possible values can be represented by $n$ Boolean variables. In either of these two examples, exactly one Boolean variable in the collection should be true at a time. A more compact representation would encode an integer value with the more standard representation with only a logarithmic number of bits.\footnote{This latter representation would, in some cases, be preferable. However, experience with constraint-solving shows that the unary representation can be more efficient when the value range is narrow.}

More generally, we translate a {\em value} of a scalar type with a collection of {\em Boolean formulas}. For example, we translate a value of a bounded integer $n..m$ to $m - n + 1$ Boolean formulas; each one of those formulas represents the truth value of the corresponding integer value.
The translation of scalar expressions is shown in Table \ref{tab:scalars}.
We have adopted the notational conventions described in Table \ref{tab:notational_conventions} to define the translations.

\begin{table}
    \centering
    \begin{tabular}{lll}
        \toprule
        Notation & Explanation     & Example\\
        \midrule
        $c$      & constant value  & $3$ \\
        $v$      & variable        & \\
        $e$      & expression      & $(v_1 \text{ add } 2) \text{ eq } v_2$ \\
        $\phi$   & Boolean formula & $(\top \wedge v_1) \vee v_2 $ \\
        $\semantics{e}{t}$ & semantic of $e$, and& $\semantics{v_1 \text{ add } 2}{0..7}$\\
                & $e$ is value of type $t$&\\
        $\tuple{e_i}_{i=1}^n$ & $\tuple{e_1,\ldots,e_n}$&\\
        $\tuple{e_1, \ldots, e_n}.i$  & picking the i-th&\\
                                      & element (i.e., $e_i$)&\\
        $|\tuple{e_1, \ldots, e_n}|$  & size of the&\\
                                      &tuple (i.e., $n$)&\\
        \bottomrule
    \end{tabular}
    \caption{Notational conventions}
    \label{tab:notational_conventions}
\end{table}

\begin{table}
    \centering
    \[ 
    \begin{array}{ll}
        \toprule
        \text{Expression}  & = \text{Translation}\\
        \midrule

        \semantics{v}{bool} & = v \\

        \semantics{c}{n..m} & = \tuple{\phi_i}_{i=n}^m, s.t.
                                  ~~~~~\phi_i = \left\{\mySubstack{
                                           \bot \text{ ~~~~if } i \neq c \\
                                           \top \text{ ~~~~if } i = c}\right. \\
        \\[-0.8em]

        \semantics{v}{n..m} & = \tuple{\semantics{v_i}{bool}}_{i=n}^m \\
        \\[-0.8em]

        \semantics{c_i}{\{c_1, \ldots, c_n\}} & = \tuple{\phi_j}_{j=1}^n, s.t.
                                  ~~~~~\phi_j = \left\{\mySubstack{
                                           \bot \text{ ~~~~if } j \neq i \\
                                           \top \text{ ~~~~if } j = i}\right. \\
        \\[-0.8em]

        \semantics{v}{\{c_1, \ldots, c_n\}} & = \tuple{\semantics{v_{c_i}}{bool}}_{i=1}^n \\
        \\[-0.8em]
        
        \semantics{\text{not } e}{bool} & = \neg \semantics{e}{bool} \\

        \semantics{e_1 \text{ and } e_2}{bool} & = \semantics{e_1}{bool} \wedge \semantics{e_2}{bool}\\

        \semantics{e_1 \text{ or } e_2}{bool} & = \semantics{e_1}{bool} \vee \semantics{e_2}{bool}\\

        \semantics{e_1 \text{ add } e_2}{n..m} & = \Tuple{\bigvee\limits_{\substack{n_1 \leq j \leq m_1,\\
                                         n_2 \leq i - j \leq m_2}}
                      \left(\substack{\semantics{e_1}{n_1..m_1}.j \wedge \\
                                      \semantics{e_2}{n_2..m_2}.(i - j)}\right)
              }_{i=n}^m\\
        \\[-0.8em]

        \semantics{e_1 \text{ sub } e_2}{n..m}& = \Tuple{\bigvee\limits_{\substack{n_1 \leq j \leq m_1,\\
                                         n_2 \leq j - i \leq m_2}}
                      \left(\substack{\semantics{e_1}{n_1..m_1}.j \wedge \\
                                      \semantics{e_2}{n_2..m_2}.(j - i)}\right)
              }_{i=n}^m\\
        \\[-0.8em]

        \semantics{e_1 \text{ mul } e_2}{n..m}& = \Tuple{\bigvee\limits_{\substack{n_1 \leq j \leq m_1,\\
                                         n_2 \leq \frac{i}{j} \leq m_2}}
                      \left(\substack{\semantics{e_1}{n_1..m_1}.j \wedge \\
                                      \semantics{e_2}{n_2..m_2}.(\frac{i}{j})}\right)
              }_{i=n}^m\\
        \\[-0.8em]

        \semantics{e_1 \text{ div } e_2}{n..m}& = \Tuple{\bigvee\limits_{\substack{n_1 \leq j \leq m_1,\\
                                         n_2 \leq \frac{j}{i} \leq m_2}}
                      \left(\substack{\semantics{e_1}{n_1..m_1}.j \wedge \\
                                      \semantics{e_2}{n_2..m_2}.(\frac{j}{i})}\right)
              }_{i=n}^m\\
        \\[-0.8em]

        \semantics{e_1 \text{ eq } e_2}{bool}& = \bigvee\limits_{j = max(n_1, n_2)}^{min(m_1, m_2)}
                      \left(\substack{\semantics{e_1}{n_1..m_1}.j \wedge \\
                                      \semantics{e_2}{n_2..m_2}.j}\right)\\
        \\[-0.8em]

        \semantics{e_1 \text{ lt } e_2}{bool}& =
        \bigvee\limits_{\substack{n_1 \leq j_1 \leq m_1,\\ j_1 < j_2 \leq m_2}}
               \left(\substack{\semantics{e_1}{n_1..m_1}.j_1 \wedge \\
                               \semantics{e_2}{n_2..m_2}.j_2}\right) \\
        \\[-0.8em]

        \semantics{e_1 \text{ leq } e_2}{bool} & =
        \semantics{e_1 \text{ lt } e_2}{bool} \vee
        \semantics{e_1 \text{ eq } e_2}{bool} \\

        \semantics{e_1 \text{ gt } e_2}{bool} & =
        \semantics{e_2 \text{ lt } e_1}{bool} \\

        \semantics{e_1 \text{ geq } e_2}{bool} & =
        \semantics{e_2 \text{ leq } e_1}{bool} \\

        \bottomrule
    \end{array}
    \]
    \caption{Translation of scalar expressions}
    \label{tab:scalars}
\end{table}

To prevent overflow in the arithmetic operations, we always cast the type of the result value to a bounded integer with the minimum range that covers all possible values. For example, if we have a value $v_1$ of type $n_1..m_1$ and another value $v_2$ of type $n_2..m_2$, then the type of expression $v_1 \text{ add } v2$ will be $n..m$, such that $n=n_1 + n_2$ and $m = m_1 + m_2$.

\subsection{Records and Unions}

Tuples are similar to records in that they consist of multiple fields of possibly different types.
Tuples are ordered, and the fields are numbered as 1,2,..., where record fields have alphanumeric names and are not a priori ordered.
Records can be trivially reduced to tuples by ordering the fields, e.g., lexicographically, and viewing $n$-field record as an $n$-tuple.

A program variable of a union type can have alternative values of different types.
For example, the values of a variable of type {\em mailing address} could be alternatively an email address (of type string) or a physical address consisting of multiple fields such as a street name, city, zip code, and country.
A union type can similarly be represented as a tuple by representing the {\em tag} (e.g., \verb+emailAddress+ and \verb+physicalAddress+ in our example) in the first component of the tuple, and then the remaining components representing the alternative values for each tag.
For example, the second component could be the string for the email address, and the third component could be a tuple for all of the physical address fields.
Often the alternative types share components of the same type.
For example, both the email address and the street name are strings, so one field could be used to store values belonging to different tags.
This helps reduce the number of bits needed to store the alternative values.

From now on, we assume that record and union types have been reduced to tuples.
We do this for the simplicity of presentation, but this would also be a good strategy for implementing these types.

Notice that {\em recursive data types}, with a field of a record or a union type pointing to another value of the same type, could not be handled by this reduction without rather strict additional restrictions, due to there being no size bounds of values of recursive data types. Typical data structures represented as recursive data types are different types of trees. For lists, another data structure involving recursion, could be given a natural Booleanized representation if limited to bounded size lists.

\subsection{Arrays, Sets, and Tuples}

Complex data types such as arrays, sets, and tuples are containers to store a collection of elements.
Each of those containers has special properties; arrays map the elements of one set to another.
Sets support set-theoretic operations over its elements.
Tuples stores elements of different types.

Above, we represented scalar values with a collection of Boolean formulas.
We can extend this definition to recursively represent the values of our complex data types.
In other words, we can represent arrays, sets, and tuples by a collection of elements, such that each one of those elements is a collection of other elements (Table \ref{tab:complex}).
By this definition, we can represent complex expressions with tree-like structures in which the leaves are Boolean formulas.

\begin{example}
\label{ex:representation}
    
The representation of a variable $v$ of the type $\tuple{0..1, \{0..2\}}\{0..1\}$ is:

\[
\begin{array}{ll}
  \multicolumn{2}{l}{\semantics{v}{\tuple{0..1, \{0..2\}}\{0..1\}}} \\

  = & \tuple{\semantics{v_0}{\tuple{0..1, \{0..2\}}},
         \semantics{v_1}{\tuple{0..1, \{0..2\}}}}\\
  \\[-1em]

  = &
  \Tuple{\mySubstack{
    \tuple{\semantics{v_{0_1}}{0..1}, \semantics{v_{0_2}}{bool\{0..2\}}}, \\
    \tuple{\semantics{v_{1_1}}{0..1}, \semantics{v_{1_2}}{bool\{0..2\}}}}} \\
  \\[-1em]

  = &
  \Tuple{\mySubstack{
    \Tuple{\tuple{v_{0_{1_0}}, v_{0_{1_1}}},
           \Tuple{\mySubstack{\semantics{v_{0_{2_0}}}{bool}, \\
                              \semantics{v_{0_{2_1}}}{bool}, \\
                              \semantics{v_{0_{2_2}}}{bool}}}}, \\
    \Tuple{\tuple{v_{1_{1_0}}, v_{1_{1_1}}},
           \Tuple{\mySubstack{\semantics{v_{1_{2_0}}}{bool}, \\
                              \semantics{v_{1_{2_1}}}{bool}, \\
                              \semantics{v_{1_{2_2}}}{bool}}}}}}\\
  \\[-1em]

  = &
  \Tuple{\mySubstack{
    \tuple{\tuple{v_{0_{1_0}}, v_{0_{1_1}}},
           \tuple{v_{0_{2_0}}, v_{0_{2_1}}, v_{0_{2_2}}}}, \\
    \tuple{\tuple{v_{1_{1_0}}, v_{1_{1_1}}},
           \tuple{v_{1_{2_0}}, v_{1_{2_1}}, v_{1_{2_2}}}}}}\\
\end{array}
\]

\end{example}

\begin{table}
    \centering
    \[
    \begin{array}{lll}
        \toprule
        \text{Expression}  && \text{Translation} \\
        \midrule
        \semantics{v}{\tuple{t_1, \ldots, t_n}} & = &
        \tuple{\semantics{v_i}{t_i}}_{i = 1}^n \\
        \\[-0.8em]

        \semantics{v}{t_v\{t_k\}} & = &
        \tuple{\semantics{v_i}{t_v}}_{i = 1}^{|\dom{t_k}|} \\
        \\[-0.8em]

        \semantics{v}{\{t\}} & = & \semantics{v}{bool\{t\}} \\

        \semantics{e.i}{t_i} & = &
        \semantics{e}{t_1 \times \cdots \times t_n}.i\\

        \semantics{e_1[e_2]}{t_v} & = &
        \fix{\semantics{e_1}{t_v\{t_k\}}}
            {\tuple{\semantics{e_2 \text{ eq } c_i)}{bool}}_{i=1}^n},\\
        &&where ~~~~~~ \dom{t_k} = \tuple{c_i}_{i=1}^n \\
        \\[-0.5em]

        \semantics{e_1 \in e_2}{bool} & = & \semantics{e_2[e_1]}{bool} \\

        \semantics{e_1 \subseteq e_2}{bool} & = &
        \bigwedge\limits_{i = 1}^{|\semantics{e_1}{\{t\}}|}
          \left(\mySubstack{\neg \semantics{e_1}{\{t\}}.i \vee \\
                            ~~~  \semantics{e_2}{\{t\}}.i} \right) \\
        \\[-0.9em]

        \semantics{e_1 \cup e_2}{\{t\}} & = &
        \tuple{\mySubstack{\semantics{e_1}{\{t\}}.i \vee \\
                           \semantics{e_2}{\{t\}}.i}}_
                               {i =  1}^{|\semantics{e_1}{\{t\}}|} \\
        \\[-0.9em]

        \semantics{e_1 \cap e_2}{\{t\}} & = &
        \tuple{\mySubstack{\semantics{e_1}{\{t\}}.i \wedge \\
                           \semantics{e_2}{\{t\}}.i}}_
                               {i =  1}^{|\semantics{e_1}{\{t\}}|} \\
        \\[-0.9em]

        \semantics{e_1 \setminus e_2}{\{t\}} & = &
        \tuple{\mySubstack{~~~  \semantics{e_1}{\{t\}}.i \wedge \\
                           \neg \semantics{e_2}{\{t\}}.i}}_
                               {i =  1}^{|\semantics{e_1}{\{t\}}|} \\
        \\[-0.7em]

        \semantics{e_1 \text { eq } e_2}{bool} & = &
        \left\{\mySubstack{\semantics{e_1}{bool} \leftrightarrow \semantics{e_2}{bool},\\
        \text{if } e_1 \text{ and } e_2 \text { are of type } bool \\
        \\
        \\
        \\
        \bigwedge\limits_{i = 1}^{|\semantics{e_1}{t}|}
        \left(\semantics{e_1}{t}.i \text{ eq } \semantics{e_2}{t}.i \right),\\
        \text{if } t \text { is not a scalar type}\\
        }\right.\\
    \end{array}
    \]
    \caption{Translation of complex data type expressions}
    \label{tab:complex}
\end{table}

\begin{table}
    \centering
    \[
    \begin{array}{lll}
        \toprule
        \text{Function}  && \text{Definition} \\
        \midrule
  
        \dom{bool} & = & \tuple{\bot, \top} \\

        \dom{n..m} & = & \tuple{i}_{i=n}^m \\

        \dom{\{e_1, \ldots, e_n\}} & = &
        \tuple{e_i}_{i=1}^n \\
        \\[-1em]

        \dom{\tuple{t_1, \ldots, t_n}} & = &
        \tuple{\tuple{c_i}_{i=1}^n}_{
            \substack{c_1 \in \dom{t_1}, \\
                      \ldots, \\
                      c_n \in \dom{t_n} \\}} \\
        \\[-0.5em]

        \dom{t_v\{t_k\}} & = & \Dom{\tuple{t_v}_{i = 1}^{|\dom{t_k}|}} \\

        \dom{\{t\}} & = & \dom{bool\{t\}} \\

        \fix{\tuple{\semantics{e_i}{bool}}_{i=1}^n}{\tuple{\phi_i}_{i=1}^n}
        & = &
        \bigvee\limits_{i=1}^n \left(\semantics{e_i}{bool} \wedge
                                     \phi_i\right) \\

        \fix{\tuple{\semantics{e_i}{t}}_{i=1}^n}
            {\tuple{\phi_i}_{i=1}^n} & = &
        \Tuple{\fix{\tuple{\semantics{e_i}{t}.j}_{i=1}^n}
                   {\tuple{\phi_i}_{i=1}^n}}_{j=1}^{|\semantics{w}{t}|} \\

        \bottomrule
    \end{array}
    \]
    \caption{Helper function definitions}
    \label{tab:helper_functions}
\end{table}

The most complicated part of the translation in Table \ref{tab:complex} is array indexing, in which we select an element of the array that is associated with the index.
To translate this expression, we construct a result value with the same form as the array elements; to fill the content of each component of this value, we iterate over the same component of the array elements to find the value of the matching index. To implement this idea, we define two helper functions in Table \ref{tab:helper_functions}.

\begin{example}
    Suppose $v$ is the variable of Example \ref{ex:representation} (a variable of type $\tuple{0..1, \{0..2\}}\{0..1\}$), and $v_k$ is a variable of type $0..1$.
    the translation of $v[v_k]$ is:

    \[
      \begin{array}{l}
        \semantics{v[v_k]}{\tuple{0..1, \{0..2\}}}\\
        \\[-0.5em]
        =\text{fix}\left(\mySubstack{\semantics{v}{\tuple{0..1, \{0..2\}}\{0..1\}},\\
                                    \tuple{\semantics{v_k \text{ eq } 0}{bool},
                                           \semantics{v_k \text{ eq } 1}{bool}}}\right)\\
        \\[-0.5em]
        =\text{fix}\left(\mySubstack{\semantics{v}{\tuple{0..1, \{0..2\}}\{0..1\}},\\
                                    \tuple{v_{k_0}, v_{k_1}}}\right)\\
        \\[-0.5em]
        =\Tuple{\text{fix}\left(\mySubstack{\semantics{v}{\tuple{0..1, \{0..2\}}\{0..1\}}.i,\\
                                    \tuple{v_{k_0}, v_{k_1}}}\right)}_{i=1}^2\\
        \\[-0.5em]
        =\Tuple{\Tuple{\text{fix}\left(\mySubstack{\semantics{v}{\tuple{0..1, \{0..2\}}\{0..1\}}.i.j,\\
                                    \tuple{v_{k_0}, v_{k_1}}}\right)}_{j=1}^2}_{i=1}^2\\
        \\[-0.5em]
        =\Tuple{\Tuple{\Tuple{\text{fix}\left(\mySubstack{\semantics{v}{\tuple{0..1, \{0..2\}}\{0..1\}}.i.j.k,\\
                                    \tuple{v_{k_0}, v_{k_1}}}\right)}_{k=1}^{|\semantics{w}{t_j}|}}_{j=1}^2}_{i=1}^2\\
        \\[-0.5em]
        =\tuple{\tuple{(v_{0_{j_k}} \wedge v_{k_0}) \vee (v_{1_{j_k}} \wedge v_{k_1})}_{k=1}^{|\semantics{w}{t_j}|}}_{j=1}^2\\
      \end{array}
    \]

    Where $t_1 = 0..1$ and $t_2 = \{0..2\}$.
\end{example}

\section{Reduction to PDDL}

So far, we have described an abstract modeling language with complex
data types; then, we devised a concrete modeling language based on this
abstract syntax to be used as a front-end of a planning system.

A natural way to implement the back-end of the planner is to map the abstract
syntax further to an existing intermediate-level modeling language such as PDDL,
that already has several scalable and robust implementations. This is what we do next.

Using the bottom-up fashion, we first explain how we construct the fundamental components of the PDDL: the parameters of actions, conditions, and actions' effects.
Since their integration into top-level components (actions, initial state, and the goal condition) is trivial, we skip explaining it; but we discuss the challenges we face, and we will provide a solution for them.

\subsection{Action Parameters}

As PDDL only supports parameterization of actions with object-valued parameters, action parameters for the extended language need to be reduced to something simpler. We use the same approach described above to represent action parameters with collections of Boolean variables.

\begin{example}
Suppose an action in the extended language has a parameter $v$ of type $\tuple{0..1, \{0..2\}}\{0..1\}$, which is represented by the following structure of Boolean variables (Example \ref{ex:representation}):
\[
  \Tuple{\mySubstack{
    \tuple{\tuple{v_{0_{1_0}}, v_{0_{1_1}}},
           \tuple{v_{0_{2_0}}, v_{0_{2_1}}, v_{0_{2_2}}}}, \\
    \tuple{\tuple{v_{1_{1_0}}, v_{1_{1_1}}},
           \tuple{v_{1_{2_0}}, v_{1_{2_1}}, v_{1_{2_2}}}}}}\\
\]

We decompose this structure to the following set of Boolean variables to determine the parameters of corresponding PDDL action.

$$\left\{v_{0_{1_0}}, v_{0_{1_1}}, v_{0_{2_0}}, v_{0_{2_1}}, v_{0_{2_2}}, v_{1_{1_0}}, v_{1_{1_1}}, v_{1_{2_0}}, v_{1_{2_1}}, v_{1_{2_2}}\right\}$$

\end{example}

\subsection{Conditions and Handling Disjunctions}

Like PDDL, the conditions (actions' preconditions, effects' conditions, and goal conditions) in our planning model are expressed by Boolean formulas.
However, Most planners do not intrinsically support disjunctions.
They compile away disjunctions by converting Boolean formulas into their disjunctive normal form (DNF) and splitting their surrounding structures (i.e., actions or effects) into multiple instances based on conjunctive terms of the DNF \cite{Helmert09}.
However, due to the exponential growth of the formula size in DNF conversion, this approach is often not feasible for the Boolean formulas generated by our translation.

Nebel \shortcite{Nebel00} has shown that general Boolean formulas in conditions cannot be reduced to conjunctions of literals without adding auxiliary actions, and the number of auxiliary actions in the plans necessarily increases super-linearly. Nebel, however, sketches reductions that increase plan length only polynomially. We use this type of reduction in our planner front-end.

More specifically, for each action $a$, we eliminate its disjunctions in multiple rounds; in round $i$, we replace each subformula of the form $\phi = \psi_1 \vee \cdots \vee \psi_n$ by $w^i_{\phi}$, such that $\psi_j, 1 \leq j \leq n,$ contains no disjunction.
After $m$ rounds that all disjunctions have been eliminated, we create $m$ auxiliary actions $b^a_1, \ldots, b^a_m$ to maintain the values of our auxiliary variables.
For each $w^i_{\phi}$, we add the set of $\{\tpl{\psi_j, w^i_{\phi} := \top} \mid 1 \leq j \leq n\}$ to the effects of the action $b^a_i$; moreover, we add $\tpl{\top, w^i_{\phi} := \bot}$ to the effects of action $a$, and initialize all auxiliary variables with the value of $\bot$ in the initial state.

It is also possible that $\psi_j, 1 \leq j \leq n,$ be a parameter of the action $a$.
In this case, we move the parameter from action $a$ to the auxiliary action $b^a_i$, and add a state variable $v^{\psi_j}$ to the problem.
Then, we replace $\psi_j$ by $v^{\psi_j}$ in $b_{i + 1}, \ldots, b_m, $ and $a$, and to make everything consistent, we add $\{\tpl{\psi_j, v^{\psi_j} := \top}, \tpl{\neg \psi_j, v^{\psi_j} := \bot}\}$ to the effects of $b^a_i$.

Finally, we enforce the sequence of $b^a_1, \ldots, b^a_n$ to be executed before the execution of action $a$, by introducing variables of $p_0, p^a_i, 1 \leq i \leq n,$ and:

\begin{itemize}
  \item adding $p_0$ and $\{\tpl{\top, p_0 := \bot}, \tpl{\top, p^a_1 := \top}\}$ to the precondition and effects of $b^a_1$, respectively,
  \item adding $p^a_{i - 1}$ and $\{\tpl{\top, p^a_{i - 1} := \bot}, \tpl{\top, p^a_i := \top}\}$ to the precondition and effects of $b^a_i, 2 \leq i \leq n$, respectively,
  \item adding $p^a_n$ and $\{\tpl{\top, p^a_n := \bot}, \tpl{\top, p_0 := \top}\}$ to the precondition and effects of action $a$, respectively, and
  \item initializing $p_0$ with $\top$ and $p^a_i, 1 \leq i \leq n$, with $\bot$ in the initial state.
\end{itemize}

\subsection{Assignments}

In our planning model, the actions' effects are described by a set of conditional assignments, which are pairs of the form $\tuple{cond, r := e}$.
Since the structures of both $\semantics{r}{t}$ and $\semantics{e}{t}$ are the same, to reduce these assignments to Boolean assignments suitable for PDDL, we can just find the Boolean variables in $\semantics{r}{t}$ and their corresponding Boolean formula in $\semantics{e}{t}$, and create conditional Boolean assignments based on them. This procedure is straightforward, except when array indexing is used to specify the reference variables.
Using the array indexing feature, we can refer to different elements of an array; the element is determined based on the values of other variables specified by the index expression.
Therefore, different Boolean variables may be selected when we have different values for the variables.

To translate the conditional assignment $\tuple{cond, r := e}$, first, we find the set of pairs of $\tpl{\semantics{v}{t}, \phi}$ for the reference expression $r$, such that $\semantics{v}{t}$ is a reference variable with the same structure as $\semantics{e}{t}$, and $\phi$ is a Boolean formula shows the condition under which $r$ indicates $v$ (Table \ref{tab:references}).
Then, we can perform the procedure described before.
More precisely, for each Boolean variable $v_b$ in $\semantics{v}{t}$, we can find the corresponding element $e_b$ in $\semantics{e}{t}$, and create a conditional Boolean assignment $\tuple{cond \wedge \phi, v_b := e_b}$.
Moreover, since PDDL only supports $\top$ and $\bot$ as the assignment values (add effects and delete effects), we further reduce it to $\tuple{cond \wedge \phi \wedge e_b, v_b := \top}$ and $\tuple{cond \wedge \phi \wedge \neg e_b, v_b := \bot}$.

\begin{table}
  \centering
  \[
  \begin{array}{lll}
    \toprule
    \text{Expression} & & \text{Definition} \\
    \midrule
    \rsemantics{v}{t} & = & \left\{\tpl{\semantics{v}{t}, \top}\right\} \\
    \\[-0.9em]

    \rsemantics{r.i}{t_i} & = &
    \left\{\tpl{\semantics{v}{t_i}.i, \phi} ~\middle|~
           \tpl{\semantics{v}{t}, \phi} \in
           \rsemantics{r}{t}\right\}, \\
    &&where, ~~ t = \tpl{t_1, \ldots, t_n} \\
    \\[-0.5em]

    \rsemantics{r[e]}{t_v} & = &
    \left\{\tpl{\semantics{v}{t_v}.i, \psi} \middle|~
           \mySubstack{\tpl{\semantics{v}{t}, \phi} \in \rsemantics{r}{t}, \\
                       1 \leq i \leq |\dom{t_k}|, \\
                       c_i = \dom{t_k}.i, \\
                       \omega = \semantics{c_i \text{ eq } e}{bool}, \\
                       \psi = \phi \wedge \omega}
           \right\},\\
    && where, ~~ t = t_v\{t_k\} \\
    \bottomrule
  \end{array}
  \]
  \caption{Finding reference variables from reference expressions}
  \label{tab:references}
\end{table}

\subsection{Action Splitting}

Most PDDL planners are based on {\em grounding}: going through all possible combinations of parameter values and creating one non-parametric action for each combination.
With complex data type parameters, grounding becomes quickly infeasible due to the high number of parameter value combinations.
For example, an action with a set-valued parameter, with values from a set with only 20 elements, would have over one million ground instances.
With 30 elements, this would be one billion.

In the disjunction elimination part, we described the idea of moving one parameter to one of its auxiliary actions, which might mitigate the issue described here by partitioning the parameters to the auxiliary action's parameters. Still, since there is no guarantee that it solves our issue entirely, or the issue might transfer from the action to its auxiliary actions, we need to solve this another way.

To approach this issue, for each action $a$, we partition its parameters into $k$ sets with at most $m$ elements.
Then, we create sub-actions $c^a_1, \ldots, c^a_k$ and move the partitioned parameters to their corresponding sub-action, precisely the same way as we described in disjunction elimination.
Furthermore, we can transfer the precondition of action $a$ to $c^a_1$, to improve the search by preventing choosing parameter values for inapplicable actions. It is worth mentioning that because we have already eliminated disjunctions, we can eliminate all parameters existing in the precondition from action $a$ by replacing them with $\top$.

This is similar to the action splitting done by \cite{ArecesBDH14} in order to be able to
ground actions with a very high number of parameter combinations.

\section{Experiments}

\begin{table}
  \centering
  \begin{tabular}{lccccc}
    \toprule
    & \# & LAMA & FDSS & MpC & FF \\
    \midrule
    Rubik    & 20 &  8 & 17 & 11 & 10\\
    Buckets  & 30 &  3 &  3 & 10 & 30\\
    Scrabble & 11 &  8 &  8 &  3 &  1\\
    \midrule
    Sudoku$_{\text{scalar}}$ & \multirow{2}{*}{46} & 3 & 8 & 19 & 4\\
    Sudoku$_{\text{array}}$  & & \textbf{19} & \textbf{25} & \textbf{46} & \textbf{16}\\
    \midrule
    Trucks$_{\text{scalar}}$ & \multirow{2}{*}{64} & 34 & 35 & \textbf{45} & \textbf{48}\\
    Trucks$_{\text{set}}$ && \textbf{42} & \textbf{38} & 16 & 38\\
    \bottomrule
  \end{tabular}
  \caption{Results of experiments}\label{ta:runtimes}
\end{table}

We pursued two goals in our experiments. The first goal was to evaluate the practicality of the proposed method to translate problems with complex data types into the most common version of PDDL that supports only Boolean variables. Moreover, our second and more specific goal was to evaluate the effectiveness and improvement of using complex data types compared to the cases that we can also describe our problems with scalar type values without too much difficulty.

For the first goal, we designed some new domains that are difficult to express with only scalar type values. These domains are: the Rubik's cube, the Buckets problems, and the Scrabble game. To describe the Rubik's cube, we need to define a three-dimensional (3D) array; its actions transform this 3D array. The Buckets problem is a famous numeric problem in which we have two buckets; we can fill up, empty, or pour water from one bucket to the other bucket as much as possible. Our goal is to reach a state that a bucket has a certain amount of water. Since our translation support bounded-range integers, we consider this problem to evaluate that feature. Finally, the Scrabble game, which is the most complicated problem, is a game to fill a board (which is a two-dimensional array) with some tiles of alphabets. In each action, a subsequent of alphabets should be chosen such that by putting them on the board, it forms a word in the dictionary. A word is an array of alphabets, and the dictionary is a set of arrays of alphabets.

The experiments are on a number of older and newer planners, including
FF \cite{HoffmannNebel01}, LAMA \cite{RichterWestphal10}, MpC \cite{Rintanen12ecaieng}, and
FDSS \cite{SeippSHH15}. We ran the planners with a 30 minute time limit, and report the number of solved instances for each planner in Table \ref{ta:runtimes}.

Our experimental results showed that the two domains of the Rubik's cube and the Buckets problem could be solved reasonably well by well-known PDDL planners. However, planners have some difficulty in solving the challenging domain of Scrabble, which means there exist plenty of potential research opportunities in this area.

To better evaluate the effects of using complex data types, we conducted other experiments to compare the performance of solving identical problems in two cases: in one case, we use complex data types, and in another case, we use only scalar data types to describe the same problem. In most existing domains, manipulating complex data types is not reasonable; scalar values provide a more simple and straightforward problem definition. On the other hand, complex data types add only complications to the problem because their operations affect a large number of elements, which makes the reasoning more challenging. However, if the intrinsic nature of a problem requires complex data types, using them to describe the problem might improve the performance.

Here, we examined two domains: Sudoku and the Trucks. In the Sudoku problem, the goal is to fill up a $9 \times 9$ board such that each digit between 1 and 9 exists in exactly one cell, in every row, column, and the $3 \times 3$ subgrids. We compared two versions of this problem; in the first version, all the board is filled at once by using array data type. In the second version, we fill the board cell-by-cell.

The second domain is the slightly modified version of the Trucks domain, used in IPC 2006. In this domain, a truck should deliver some packages to some location by loading them from other locations and transferring them to their destinations. However, there are also some time constraints that specify the latest arrival time of the packages. In our version, the truck can load a set of packages at once, so we no longer have the spatial constraints of the original domain. However, we enforce the capacity constraints in another way, such that the truck could not load all packages at once.

The results show that using the array data type in the Sudoku domain significantly improves the performance compared to the version with only scalar values. This is mainly because in the former version the action specifies all the required constraints at the begining, compared to the latter version that dead-end nodes will be determined after performing some actions.

Although in the Trucks domain, some planners performed slightly better in the version with the set data type, the results show that Madagascar's performance has been considerably worsen in this version.
This is because Madagascar often critically relies on the possibility of performing multiple actions in parallel, and this is not allowed by the way our auxiliary actions for parameterization and elimination of disjunctions are constructed.

\section{Conclusion}

We have proposed a very expressive modeling language for planning, with
a rich collection of data types, and devised a translation of this language
first to Boolean logic, and then further to the Planning Domain Description
Language PDDL. We have formalized in our modeling language a number of
planning problems which would be clumsy to write in PDDL directly, and shown
that they can be solved with off-the-shelf domain-independent planners for PDDL.

We also demonstrated, very surprisingly, that in one case a handcrafted PDDL
formalization is solved by PDDL planners {\em less} efficiently than the
generated PDDL formalization produced automatically from our higher-level
formalization.

Although adapting some well-known planning techniques to more expressive
modeling languages is often a challenge, the potential of the compact
problem representations has been recognized and it has already led to
successes, for example in the case of Functional STRIPS \cite{FrancesGeffner16}.
Important part of future work is
investigating the powerful search methods that directly work on the more
expressive language, rather than going through a less powerful intermediate
language such as PDDL, as we have done in this work.

\bibliography{references}

\end{document}